\documentclass[summary]{ursi}

\usepackage[
style=numeric-comp,
sorting=none
]{biblatex}
\usepackage[T1]{fontenc}
\usepackage{newtxtext,newtxmath}
\usepackage{xcolor}
\usepackage{svg}
\usepackage{float}

\svgsetup{inkscapelatex=false}
\usepackage[utf8]{inputenc}

\usepackage{tikz}
\usepackage{pgfplots}
\usepackage{pgfplotstable}
\pgfplotsset{compat=1.7}
\usepackage{subcaption}
\usepgfplotslibrary{groupplots}
\usepackage{booktabs}

\title{Deep supervised hashing for fast retrieval of radio image cubes}

\author{Steven Ndung'u\affref{ref1}, Trienko Grobler\affref{ref2},
   Stefan J. Wijnholds\affref{ref2}\affref{ref3},
   Dimka Karastoyanova \affref{ref1}
   and George Azzopardi\affref{ref1}
   }

\affiliation{%
  \aff{ref1}{University of Groningen, Groningen, The Netherlands; e-mail: s.n.machetho@rug.nl}
  \aff{ref2}{University of Stellenbosch, Cape Town, South africa; e-mail: tlgrobler@sun.ac.za}
  \aff{ref3}{ASTRON, Dwingeloo, The Netherlands; e-mail: wijnholds@astron.nl}
}

\begin{document}

\maketitle

\begin{abstract}

The shear number of sources that will be detected by next-generation radio surveys will be astronomical, which will result in serendipitous discoveries. Data-dependent deep hashing algorithms have been shown to be efficient at image retrieval tasks in the fields of computer vision and multimedia. However, there are limited applications of these methodologies in the field of astronomy. In this work, we utilize deep hashing to rapidly search for similar images in a large database. The experiment uses a balanced dataset of 2708 samples consisting of four classes: Compact, FRI, FRII, and Bent. The performance of the method was evaluated using the mean average precision (mAP) metric where a precision of 88.5\% was achieved. The experimental results demonstrate the capability to search and retrieve similar radio images efficiently and at scale. The retrieval is based on the Hamming distance between the binary hash of the query image and those of the reference images in the database.
  
\end{abstract}

\section{Introduction}
\label{introduction}

In recent years, radio astronomy has experienced exponential data growth: next-generation radio surveys are producing massive data sets with tens of millions of unknown radio sources \cite{norris2017extragalactic}. With the exponential increase of high-resolution radio images, management and optimal exploitation of the desired galaxies have become challenging tasks. Thus, there is a growing need for fast, efficient and effective image retrieval for a given query galaxy. Image retrieval/indexing of radio galaxies is the process of finding and identifying galaxies with similar morphological structures in a large database of galaxies, similar to a query image. Efficient retrieval of similar radio galaxies enables astronomers to categorize and study their evolution and discover rare astronomical objects and phenomena. 

There has been gradual advancement in astronomy, with focus on image retrieval research. For instance, content-based image retrieval (CBIR) \cite{Csillaghy_2000,AbdElAziz2017AutomaticDO}, and text-based image retrieval (TBIR) \cite{walmsley2022practical} approaches  have been applied in finding the `nearest neighbour' search objects from a reference database in astronomy. To the best of our knowledge, image retrieval is, however, hardly addressed in \textit{radio} astronomy. The goal of this work is to demonstrate how image retrieval can be effectively applied in radio astronomy.

 \begin{figure}[t]
\footnotesize
\centering
\includegraphics[scale=0.7]{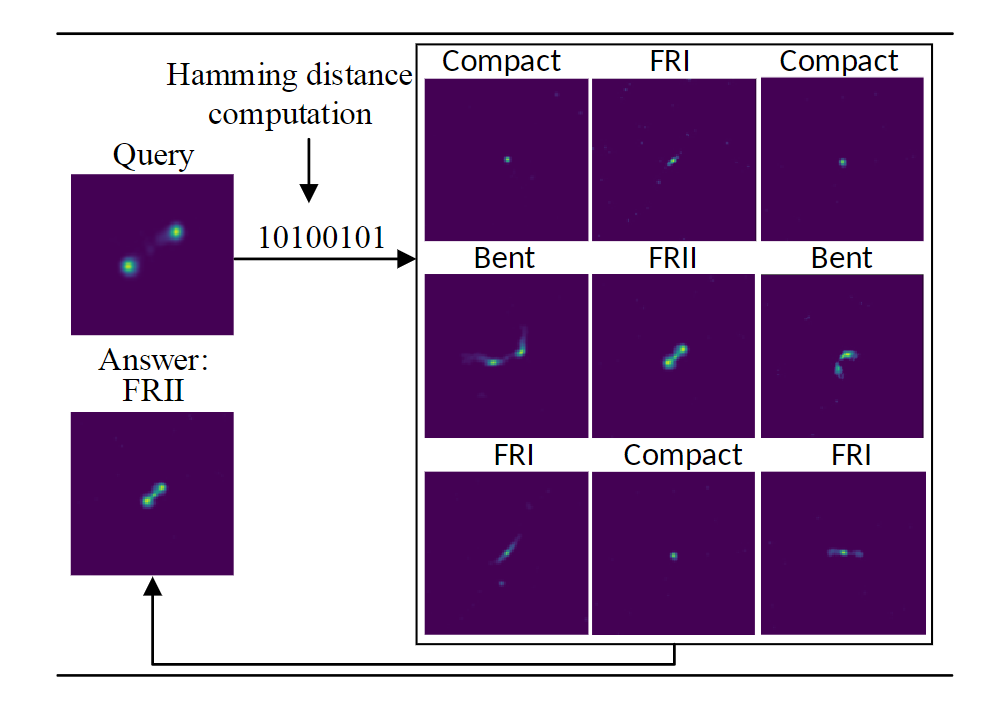}

 \caption{A schematic overview of the image retrieval process based on a query image. The Hamming distance is calculated between the query image and every reference image. The answer image represents the top 1 image retrieved.}
\label{fig:intro_image_retrieval_process}
\end{figure}

Data-dependent deep hashing algorithms utilizing convolutional neural networks (CNNs) are widely studied in image retrieval problems in the fields of multimedia and computer vision  \cite{liu2016deep,dubey2022vision, nayak2022design}. This is attributed to the ability of the deep hashing CNN to extract and learn unique image signatures \cite{lin2015deep}. The algorithms are used to create compact and low-dimensional binary representations of both the query image and the database of images. Then, a similarity calculation is performed between the binary representation of the query image and each binary representation of the database images using the Hamming distance.  The resultant distance values are then used to rank the images (e.g. the top 100) being retrieved. Fig.~\ref{fig:intro_image_retrieval_process} summarises this process.

\section{Data and Methodology}
\subsection{Dataset}
The data sample of radio galaxies used in this paper was constructed using information from multiple catalogs as compiled and processed by Samudre et al.~\cite{samudre2022data}. The original dataset is composed of four classes of galaxies: Compact (406 samples), FRI (389 samples), FRII (679 samples), and Bent (508 samples), which are distributed among the given training, validation, and test datasets \cite{gendre2008combined,10.1111/j.1365-2966.2010.16413.x,proctor2011morphological,capetti2017fricat,capetti2017friicat,baldi2018fr0cat} (Table~\ref{original_dataset_distrb}). This dataset was processed by Samudre et al.~\cite{samudre2022data} in a similar manner to the steps described in \cite{Aniyan_2017}. Furthermore, Samudre et al.~\cite{samudre2022data} balanced the train and validation datasets by upsampling the underrepresented classes through randomly duplicating samples from the original dataset. As a result, a balanced dataset of 2708 samples was obtained, composed of Compact (675 samples), FRI (674 samples), FRII (679 samples), and Bent (680 samples). 

In this work we also reshape the images to $224\times224$ pixels as an additional preprocessing step to all images during the training and application of the model.
\begin{table}
\caption{\textbf{The distribution of the original dataset spread across the  training, validation and testing images.}}
\label{original_dataset_distrb}
\begin{tabular}{lcccc}
\toprule
\textbf{Type} & \textbf{Sample size} & \textbf{Training} & \textbf{Validation} & \textbf{Test} \\
\midrule Bent & 508 & 305 & 100 & 103 \\
Compact & 406 & 226 & 80 & 100 \\
FRI & 389 & 215 & 74 & 100 \\
FRII & 679 & 434 & 144 & 101 \\
\midrule Total & 1982 & 1180 & 398 & 404 \\
\bottomrule
\end{tabular}
\end{table}

\subsection{Method}

\begin{figure*}[t]
\footnotesize
\centering
\includegraphics[scale=0.7]{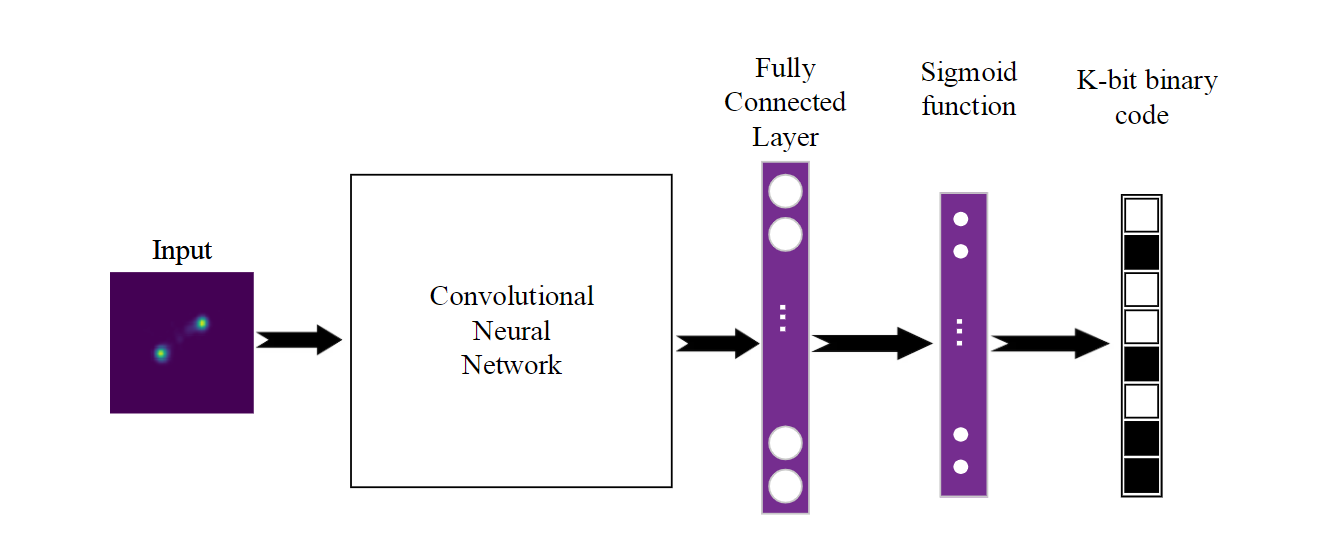}

 \caption{Structural schematic of a CNN architecture that generates a hash code to a given image.}
\label{fig:dcnn_nerwork}
\end{figure*}

The proposed image retrieval framework consists of three main phases: training, encoding and image retrieval. The first phase of the method involves pre-training a CNN on a large dataset such as ImageNet \cite{deng2009imagenet} to learn important features, patterns and representations of the images.  It is then fine-tuned with a dataset specific to the target domain (radio astronomical images). This allows the deep CNN to understand the critical morphological features on the images to map them onto a unique image representation \cite{lin2015deep}. Notably, the final layer in the proposed framework is a fully connected layer, and its neuron activations are regulated by the preceding layers, which encode semantics, to achieve the hashing process. The sigmoid activation function is used after the last layer such that all values are forced to lie in the range [0,1]. Fig.~\ref{fig:dcnn_nerwork} shows the structural schematic structure of the deep hashing model. 

In the second phase (encoding phase), the trained model is used to generate a real-valued vector for each of the training (reference) images. Such vectors are then binarized by a thresholding operation that transforms to 1 all values above a given threshold, and 0 otherwise. The same encoding operation is also applied to a given query image. These binary image representations are compact and efficient, which are then used for image retrieval. The idea behind this approach is that images with similar morphological structures should have similar binary encodings.

 Finally, the third phase is image retrieval, where binary image representations are used to match similar images. We use the Hamming distance to compare the binary representation of a query image from the test image database with the hash codes of the images in the training (reference) database. The reference images are then sorted and ranked based on the Hamming distances. Images with the shortest distance to the query image are considered to be the most similar. From the list of ranked images, the top 100 images are retrieved.

\subsection{Evaluation Metrics}
Similar to previous studies, we assess the performance of our deep CNN algorithm following the widely adopted mean average precision (mAP) metric \cite{turpin2006user}:

\begin{equation}
\text{AP} = \frac{1}{\text{GTP} }\sum_{i=1}^{n}\text{Precision}(i)\times\text{Rel}(i),
\end{equation}

\begin{equation}
\text{mAP} = \frac{1}{N_q }\sum_{j=1}^{N_q}AP_j,
\end{equation}

where AP represents the average precision of one query, with $n$ being the total number of reference images, and $\text{GTP}$ the total number of ground truth positives, $\text{Precision}(i)$ is the precision of the top $i$ ranked reference images and $\text{Rel}(i)$ is an indicator variable that is 1 if the $i$th image is relevant and 0 otherwise. Finally, the mAP is computed as the average of all AP values obtained for all $N_q$ query images.

\section{Experimental Results and Discussion}

Our model is constructed using a transfer learning paradigm. We fine-tune a DenseNet161  \cite{huang2017densely}, which is pre-trained on ImageNet \cite{deng2009imagenet}. The fine-tuning of the DenseNet161 model consisted of two steps. First, all the layers of the pre-trained model are frozen except for the final, customized layer. This layer is then trained for 15 epochs on the radio data. In the second step, the model is unfrozen and fine-tuned for 200 iterations using an optimal learning rate obtained through the cyclical learning rates approach \cite{smith2017cyclical}. The learning rate is bounded within the range of 0 to 10$^{-3}$ for fast model convergence and weight decay of 10$^{-3}$ for model regularization. We designed our model such that the final output layer generates an eight-element hash code. For the model to learn discriminative features that preserve the similarity of the images, we used the triplet margin loss function \cite{BMVC2016_119}. The loss function is designed such that the outputs of similar images are pulled together while pushing away dissimilar images. Therefore, the model effectively learns the semantic structure of similar images.

\begin{figure}[t]
\footnotesize
\centering
\includegraphics[scale=0.65]{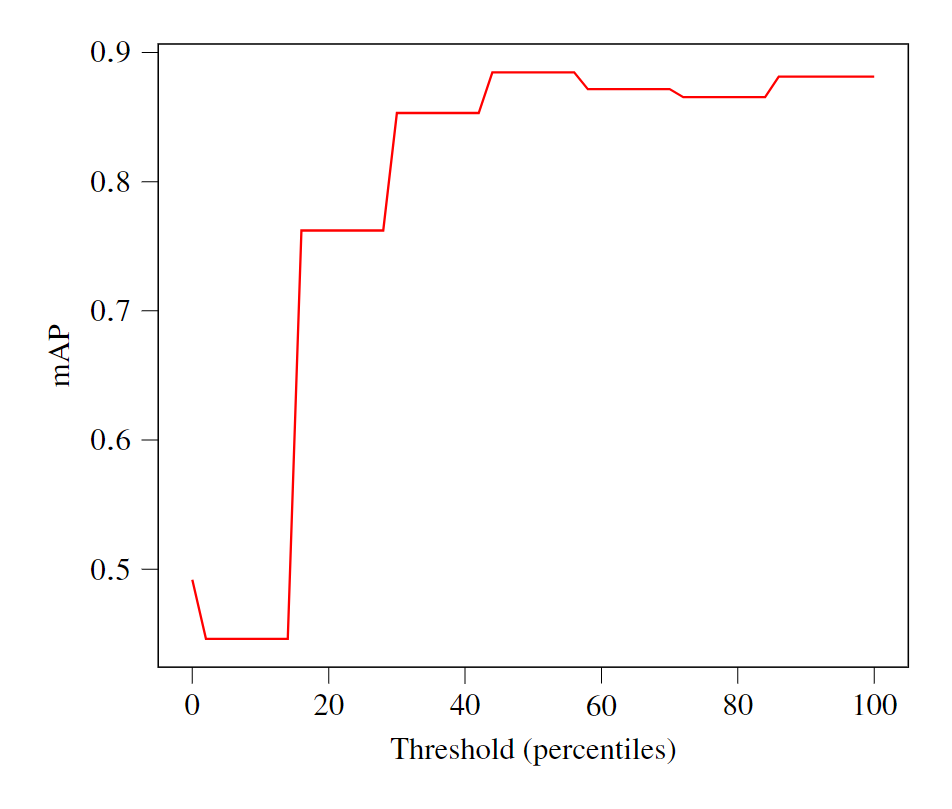}
\caption{The mAP as a function of the threshold applied in the encoding phase.}
\label{fig:map_threshold_curve}
\end{figure}

\begin{figure*}[t]
\footnotesize
\centering
\includegraphics[scale=0.65]{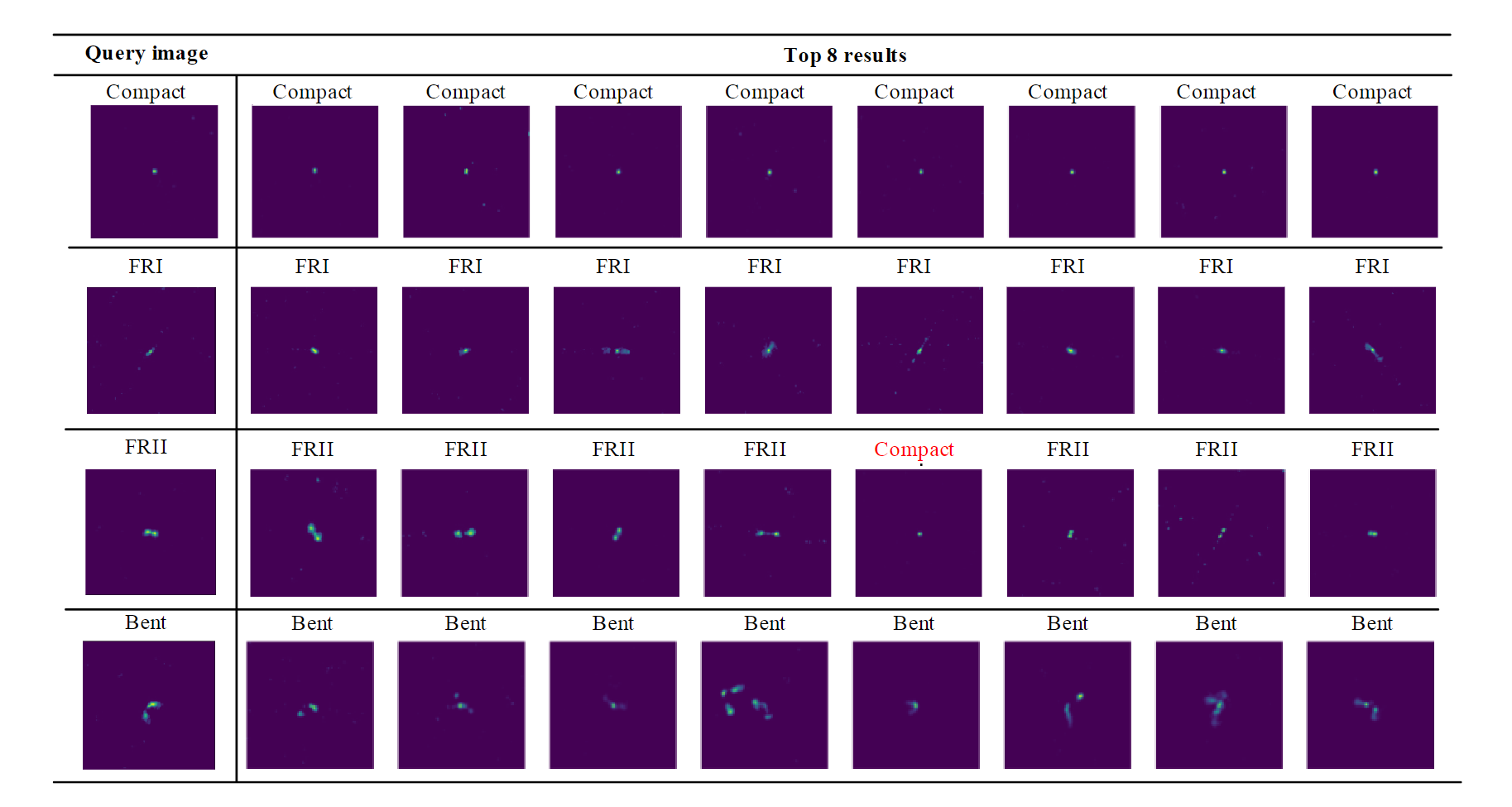}

 \caption{Example of 4 query images and the respective top-ranked 8 retrieved images. The label on each image is the true label. The red label in the 3rd row indicates a false positive.}
\label{fig:Image_Retrieval}
\end{figure*}

We evaluate the proposed framework on the given test set with different threshold percentile values used in the encoding phase and show the results in Fig.~\ref{fig:map_threshold_curve}. For instance, a threshold percentile of 50\% for a given query or reference image means that all values equal or above the median of its 8-element vector are encoded as 1 bits, otherwise 0. The maximum mAP of 0.885 is achieved at 44$^{th}$ to 56$^{th}$ percentiles (Fig.~\ref{fig:map_threshold_curve}). In literature, researchers evaluate the quality of image retrieval algorithms with mAP between 55\% and 95\% \cite{turpin2006user} considered to be good. This implies that our model's performance is stable and  reliable. Moreover, this can be affirmed by observing the top images retrieved in a given query. In Fig.~\ref{fig:Image_Retrieval}, it is clear that most of the retrieved images belong to the query image class. Moreover, the images that do not belong to the class of the query image have a very high resemblance to the query image. 

We believe that the application of deep hashing algorithms in radio astronomy holds great potential in serendipitous discoveries owing to the massive datasets generated by modern telescopes. In future, we will investigate further enhancement and efficient strategies, such as the impact of varying the length of the binary representations of the images.

\section{Conclusion}

We presented a deep CNN model based on transfer learning for image retrieval in radio astronomy. The model effectively learns discriminative features, resulting in efficient and effective retrieval of similar (morphologically) radio galaxies, with a mean average precision of 88.5\%.

The key advantages of using this technique in radio astronomy are: the learning of compact representations, speed, scalability, robustness to noise (since the underlying features are learned from the images, rather than the exact pixel values) and improved retrieval performance. Therefore, this strategy is applicable to large-scale radio astronomical data generated by modern telescopes, helping astronomers quickly locate and study interesting celestial objects.

\section*{Acknowledgements}

Part of this work is supported by the Foundation for Dutch Scientific Research Institutes. 


\printbibliography

\end{document}